\documentclass[10pt,twocolumn,letterpaper]{article}

\usepackage[camera-ready]{cvpr}      

\usepackage{graphicx}
\usepackage{amsmath}
\usepackage{amssymb}
\usepackage{booktabs}
\usepackage{bm}
\usepackage{mathtools}
\usepackage{multirow}
\usepackage{pifont}
\newcommand{\norm}[1]{\left\lVert#1\right\rVert}
\newcommand{\cmark}{\text{\ding{51}}}
\newcommand{\xmark}{\text{\ding{55}}}
\usepackage[accsupp]{axessibility}
\usepackage{gensymb}
\usepackage[pagebackref,breaklinks,colorlinks]{hyperref}

\usepackage[capitalize]{cleveref}
\crefname{section}{Sec.}{Secs.}
\Crefname{section}{Section}{Sections}
\Crefname{table}{Table}{Tables}
\crefname{table}{Tab.}{Tabs.}

\def\cvprPaperID{5258} 
\def\confName{CVPR}
\def\confYear{2022}

\begin{document}

\title{Fourier Document Restoration for \\ 
Robust Document Dewarping and Recognition}

\author{Chuhui Xue$^1$, \:Zichen Tian$^1$, \:Fangneng Zhan$^1$, \:Shijian Lu$^1$, \:Song Bai$^2$\\
$^1$Nanyang Technological University, $^2$ByteDance\\
{\tt\small xuec0003@e.ntu.edu.sg, \{zichen.tian,shijian.lu,fnzhan\}@ntu.edu.sg, songbai.site@gmail.com}
}

\twocolumn[{%
\renewcommand\twocolumn[1][]{#1}%
\maketitle
\begin{center}
    \centering
    \captionsetup{type=figure}
    \includegraphics[width=\linewidth]{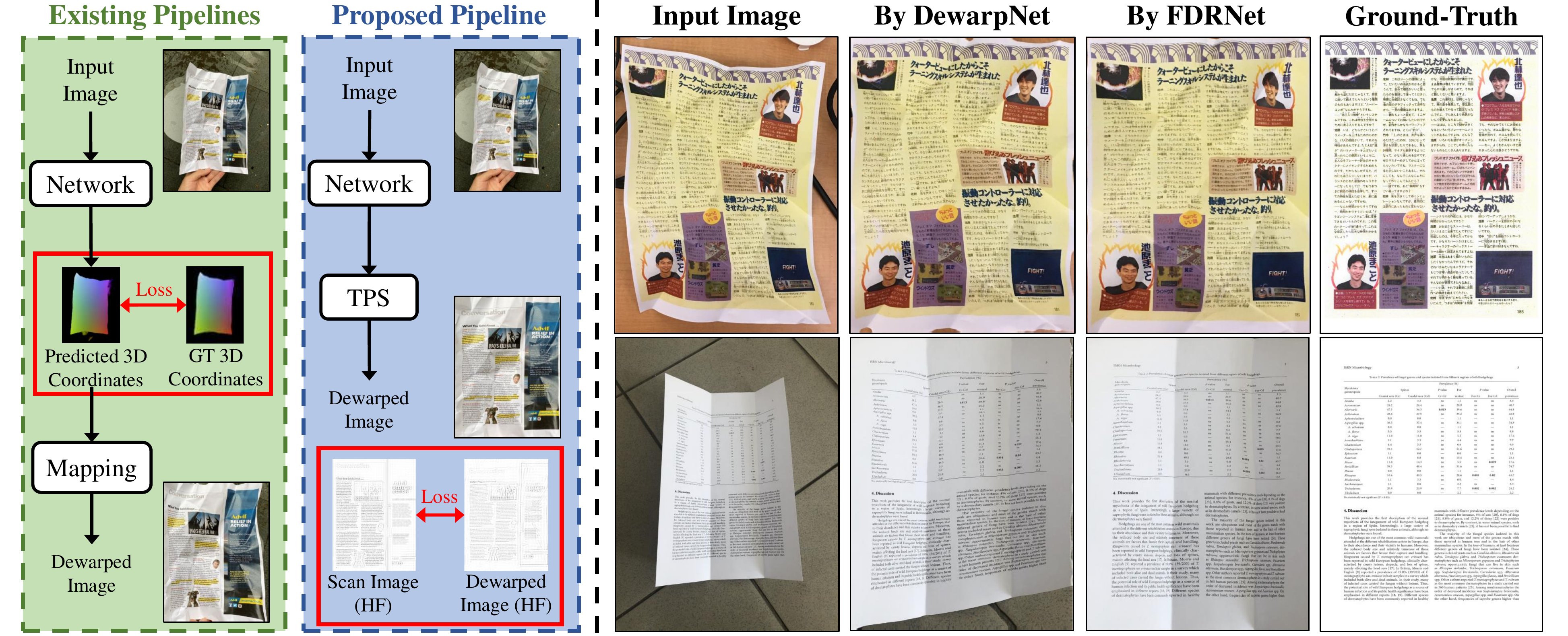}
    \captionof{figure}{\label{fig:intro}
    \textbf{Existing document dewarping and the proposed FDRNet:} Existing document dewarping learns to predict 3D coordinates of each pixel in camera document for dewarping, which often struggle when handling documents with irregular distortions or large depth variations as shown in column 2 in the right graph. FDRNet instead focuses on the high-frequency components of document contents and learns to dewarp the whole document with Thin-Plate Spline (TPS) transformation. It is robust to irregular deformations and depth variations as shown in column 3 in the right graph, and requires much less simply annotated training data.
    }
\end{center}%
}]

\begin{abstract}

State-of-the-art document dewarping techniques learn to predict 3-dimensional information of documents which are prone to errors while dealing with documents with irregular distortions or large variations in depth. This paper presents FDRNet, a Fourier Document Restoration Network that can restore documents with different distortions and improve document recognition in a reliable and simpler manner. FDRNet focuses on high-frequency components in the Fourier space that capture most structural information but are largely free of degradation in appearance. It dewarps documents by a flexible Thin-Plate Spline transformation which can handle various deformations effectively without requiring deformation annotations in training. These features allow FDRNet to learn from a small amount of simply labeled training images, and the learned model can dewarp documents with complex geometric distortion and recognize the restored texts accurately. To facilitate document restoration research, we create a benchmark dataset consisting of over one thousand camera documents with different types of geometric and photometric distortion. Extensive experiments show that FDRNet outperforms the state-of-the-art by large margins on both dewarping and text recognition tasks. In addition, FDRNet requires a small amount of simply labeled training data and is easy to deploy. The proposed dataset is available at \url{https://sg-vilab.github.io/event/warpdoc/}.

\end{abstract}



\section{Introduction}
\label{sec:intro}
Automated document recognition is critical in many applications such as library digitization, office automation, e-business, etc. It has been well solved by optical character recognition (OCR) technology if documents are properly scanned by document scanners. But for increasing document images captured by various camera sensors, OCR software often encounters various recognition problems due to two major factors. First, document texts captured by cameras often lie over a curved or folded surface and suffer from different types of geometric distortions such as document warping, folding, and perspective views as illustrated in \cref{fig:intro}. Second, document texts captured by cameras often suffer from different types of photometric distortion due to uneven illuminations, motion, shadows, etc. Accurate recognition of document texts captured by camera sensors remains a grand challenge in the document analysis and recognition research community.

Document restoration has been investigated extensively for better recognition of documents captured by various camera sensors. Recent data-driven methods \cite{feng2021doctr,das2019dewarpnet} synthesize 3D document images with various distortions and learn document distortions by predicting the 3D coordinates of \textit{each pixel} in warped documents which have achieved very impressive performances on document dewarping task. However, these methods are facing three challenges.
First, most pixels in document images suffer from regular distortions of perspective or curvature, whereas only a small portion of pixels exhibit irregular deformations (e.g. pixels around crumples). Such pixel-level data imbalance often leads to degraded performance for existing pixel-level regression-based models while handling documents with irregular deformations as shown in the first row on the right of \cref{fig:intro}. 
Second, most existing document dewarping methods perform poorly when documents are far away from the camera as illustrated in the second row on the right of \cref{fig:intro}. This is largely because existing methods often struggle in predicting document 3D coordinates when the document depth has large variations.
Third, most existing models are trained on large amounts of synthetic images, where the synthesis is complicated requiring to collect 3D coordinates by special hardware (i.e. depth camera) and a large number of scanned document images (i.e. 100,000 synthetic images from 3D coordinates of 1,000 documents and 7,200 scanned images in \cite{das2019dewarpnet}). This makes it challenging to generalize existing methods to new tasks and domains. 

We design FDRNet, an end-to-end trainable document restoration network that focuses on document contents and aims for better document recognition. FDRNet is inspired by the observation that geometric distortions in document images can be largely inferred from high-frequency components in Fourier space whereas appearance degradation is largely encoded in low-frequency components. Document restoration and recognition should therefore focus on high-frequency components capturing document structures and contents and ignoring interfering low-frequency components capturing largely appearance noises. We thus design FDRNet to learn geometric distortions by focusing on high-frequency information of the whole document instead of 3D coordinates of each pixel which helps tackle the challenge of pixel-level data imbalance and document depth variation effectively. FDRNet is powered by Thin-Plate Spline transformation which helps not only reduce training data significantly but also eliminate the need for 3D document coordinates ground-truthing and the complex data collection process in training. Furthermore, we introduce WarpDoc, a benchmarking dataset with more than one thousand document images with different types of degradation in geometry and appearance that greatly help for better validation of document dewarping models. Extensive experiments show that FDRNet achieves superior document restoration as illustrated in \cref{fig:intro}. 

The contributions of this work are three-fold. First, we design FDRNet, an end-to-end trainable document restoration network that can remove geometric and appearance degradation from camera images of documents and improve document recognition significantly. Second, FDRNet handles document restoration and recognition by focusing on high-frequency components in the Fourier space which helps reduce training data and improve model generalization and usability greatly. Third, we create a dataset with more than one thousand camera images of documents which is very valuable to future research in the restoration and recognition of documents captured by cameras.

\begin{figure*}[!ht]
  \centering
  \includegraphics[width=\linewidth]{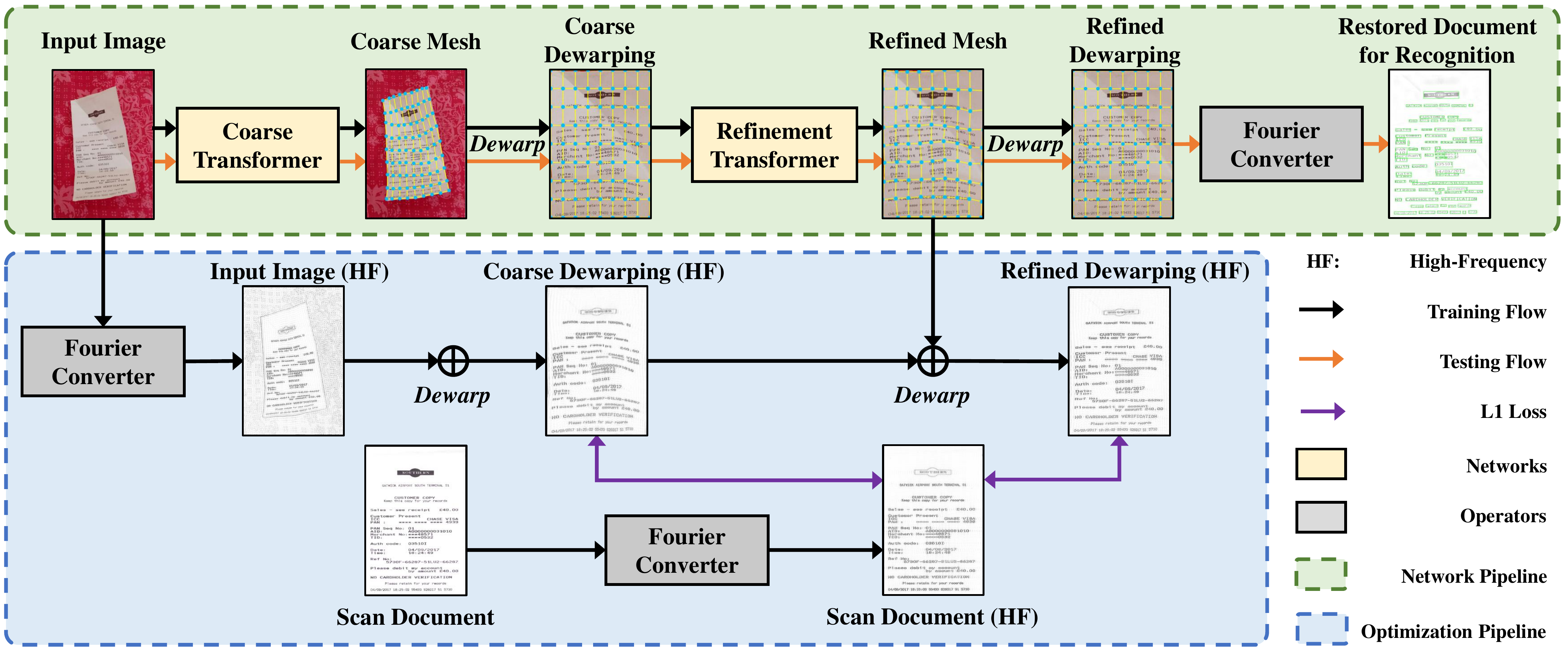}
\caption{
\textbf{The framework of the proposed FDRNet}: With camera captured \textit{Input Image}, FDRNet learns to predict control points (for thin-plate-spline transformation in dewarping) by a \textit{Coarse Transformer} and a \textit{Refinement Transformer} and use the predicted control points as the node of \textit{Coarse Mesh} and \textit{Refined Mesh} for document dewarping. It computes rectification losses ($L_1$ loss) over high-frequency information of the \textit{Input Images} as produced by a \textit{Fourier Converter} and the corresponding \textit{Scanned Documents} without using any annotations in training. In inference, the dewarped document is fed to a \textit{Fourier Converter} for photometric restoration and recognition.
}
\label{fig:overview}
\end{figure*}

\section{Related Works}
\label{sec:related_works}
\subsection{Geometric Document Restoration}
Document texts captured by camera sensors often lie over a curved/folded surface and suffer from various perspective distortions that hinder document recognition significantly. Document dewarping has been studied extensively for flattening documents into a recognition-friendly form. Traditional methods dewarp documents by reconstructing 3D document shapes \cite{brown2001document, zhang2008improved, meng2014active, ulges2004document, you2017multiview, yamashita2004shape, koo2009composition, ostlund2012laplacian, meng2018exploiting, tian2011rectification} or extracting 2D image features \cite{tsoi2007multi, wada1997shape, courteille2007shape, zhang2009unified, ezaki2005dewarping, fu2007model, liu2015restoring, lu2006partition, stamatopoulos2010goal, liang2008geometric, ulges2005document, meng2011metric,lu2006document,lu2006restoration}. On the other hand, extracting 2D features often involves various heuristic parameters and 3D reconstruction is complicated and sensitive to various noises. In recent years, some work \cite{das2017common, ma2018docunet, das2019dewarpnet} exploits deep neural networks to learn document shapes from 2D/3D synthetic document images. However, such data-driven methods require a large amount of synthetic data that are complicated and time-consuming to collect.

Our proposed FDRNet dewarps documents by learning 2D deep network features with little heuristics. Instead of using large amounts of synthetic data \cite{das2017common, ma2018docunet, das2019dewarpnet, feng2021doctr}, it learns from high-frequency components of real document images which allows learning superior geometric document restoration models with a small amount of training data.

\subsection{Photometric Document Restoration}

Document images captured by camera devices often suffer from various illumination noises such as occlusion shadows as induced by photographers or documents themselves. Such illumination noises complicate text segmentation from document background, which could degrade text recognition performance significantly. Different photometric restoration and document image binarization techniques \cite{lu2012binarization, lu2010binarization,lins2019icdar,almeida2018new,lins2017binarizing} have been reported for segmenting texts from various unevenly illuminated document images. On the other hand, most existing works are either computationally intensive \cite{lu2012binarization,da2014later} or sensitive to heuristic parameters \cite{lu2010binarization,almeida2018new,su2010binarization,lu2007binarization,bolan2010self} and not a good fit as a pre-processing step of document recognition. More recently, some approaches \cite{li2019document,feng2021doctr} correct illuminations of documents by patch-based networks. Our proposed technique handles illumination noises by extracting high-frequency document information, which is efficient and robust and involves minimal heuristics.

\section{Methodology}

The proposed FDRNet consists of three components including a \textit{Coarse Transformer}, a \textit{Refinement Transformer}, and a \textit{Fourier Converter} as illustrated in \cref{fig:overview}. The \textit{Coarse Transformer} and \textit{Refinement Transformer} learn to dewarp documents in a coarse-to-fine manner. The \textit{Fourier Converter} extracts high-frequency information of document images for effective and efficient network training as shown in the \textit{Optimization Pipeline} as highlighted in green in \cref{fig:overview}. Additionally, it extracts high-frequency content information for better document recognition as shown at the right end of the \textit{Network Pipeline} as highlighted in blue in \cref{fig:overview}.

\subsection{Coarse-To-Fine Transformer}\label{sec:net}
FDRNet dewarps document images in a coarse-to-fine manner by using a Coarse Transformer and a Refinement Transformer. The two transformers share the same architecture Spatial Transformer Network (STN) \cite{jaderberg2015spatial} that models the spatial transformation as learnable networks. Specifically, the Coarse Transformer learns to localize the document region in the input image and dewarps the located document region coarsely. The Refinement Transformer takes the dewarped document image from the Coarse Transformer and improves the dewarping further. 

We adopt Thin-Plate-Spline \cite{bookstein1989principal} (TPS) as the spatial transformation in document dewarping. TPS transformation is determined by two sets of control points with a one-to-one correspondence between a pair of warped and flat document images, and it computes a spatial deformation function for every control point to predict geometric distortions. In FDRNet, we define the control points as mesh grid and the network learns to predict the mesh grid of document region in the input image (i.e. the blue dots in \textit{Predicted Mesh} in \cref{fig:overview}). With the predicted mesh grid, TPS transforms them to the regular mesh grid (i.e. the blue dots in \textit{Coarse Dewarping} and \textit{Refined Dewarping} in \cref{fig:overview}) to achieve document dewarping. The mesh grid can have different sizes and our study shows that a $9 \times 9$ mesh grid (with 81 control points) is sufficient for document dewarping.

By denoting the predicted mesh grid points by $P=[\bm{t_1}, \bm{t_2}, ..., \bm{t_k}]^T$ and the regular mesh grid points by $P'=[\bm{t'_1}, \bm{t'_2}, ..., \bm{t'_k}]^T$, the TPS transformation parameters can be determined as follows:
\begin{equation}
C_x={
\begin{bmatrix}
S & 1_k & P\\ 
1_{k}^{T} & 0 & 0 \\ 
P^T & 0 & 0
\end{bmatrix}
}^{-1} \cdot  
\begin{bmatrix}
P_x'\\ 
0\\ 
0
\end{bmatrix} ,
\label{eq1}
\end{equation} 
where each element $(S)_{ij}$ in $S$ is determined by $\phi(\bm{t_i} - \bm{t_j})$ and $\phi(\bm{r})$ is defined by $\norm{\bm{r}}^2_2log\norm{\bm{r}}^2_2$. $P'_x$ refer to $x$ coordinates of $P'$. Similarly, $C_y$ can be obtained by replacing $P'_x$ by $P'_y$. Hence, we can get $C = [C_x, C_y]$. Finally, for each control point of the document region in the input image $\bm{u}$, the corresponding point $\bm{u'}$ in the dewarped document can be determined by:
\begin{equation}
\bm{u}' = C \cdot \bm{u} .
\label{eq2}
\end{equation}

\begin{figure}[!t]
  \centering
  \includegraphics[width=\linewidth]{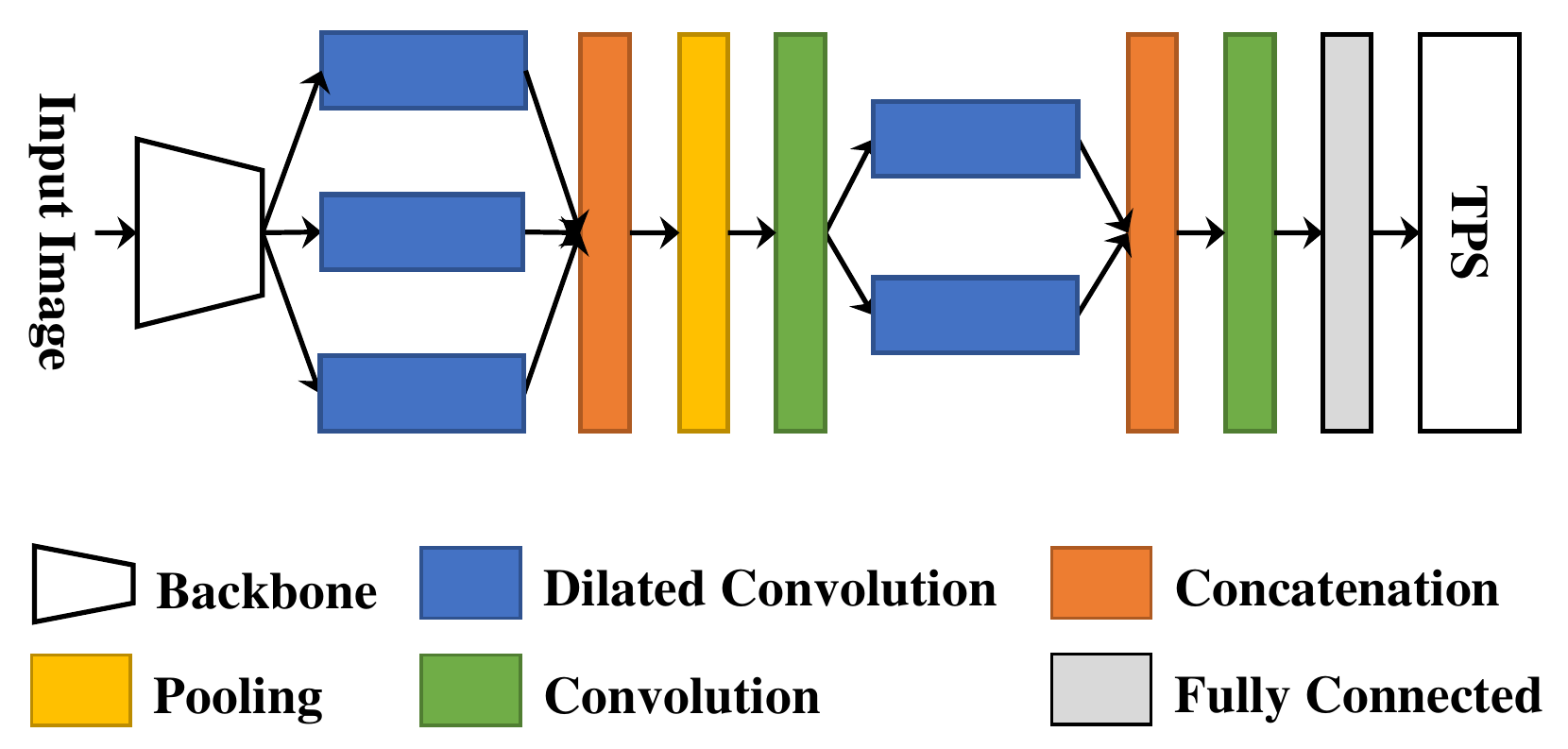}
\caption{
The \textbf{Architecture} of \textit{Coarse Transformer} and \textit{Refinement Transformer} for coarse-to-fine document distortion estimation and rectification (TPS: Thin-Plate-Spline)
}
\label{fig:net}
\end{figure}

Note that the predicted mesh grid points are initialized by regular mesh grid in the implementation. Since all operators in the TPS transformation are differentiable, the \textit{Coarse Transformer} and \textit{Refinement Transformer} can learn to localize document mesh grid points by gradient backpropagation without requiring any annotation of document mesh grids. Additionally, we adopt stacked dilated convolution \cite{schuster2019sdc,yu2015multi} in the two transformers to enlarge the network receptive field since the mesh grid localization requires to focus on high-level document content information. \cref{fig:net} shows the detailed structure of the Coarse and Refinement Transformers. Specifically, document features are first extracted by a backbone network which are then fed to three stacked dilated convolutional layers followed by two stacked dilated convolutional layers of different dilation rates \cite{wang2018understanding}. The network finally predicts a set of control points (as the document mesh grid as illustrated in \textit{Predicted Mesh} in \cref{fig:overview}) and passes them to TPS for document dewarping.

\begin{figure}[!t]
  \centering
  \includegraphics[width=\linewidth]{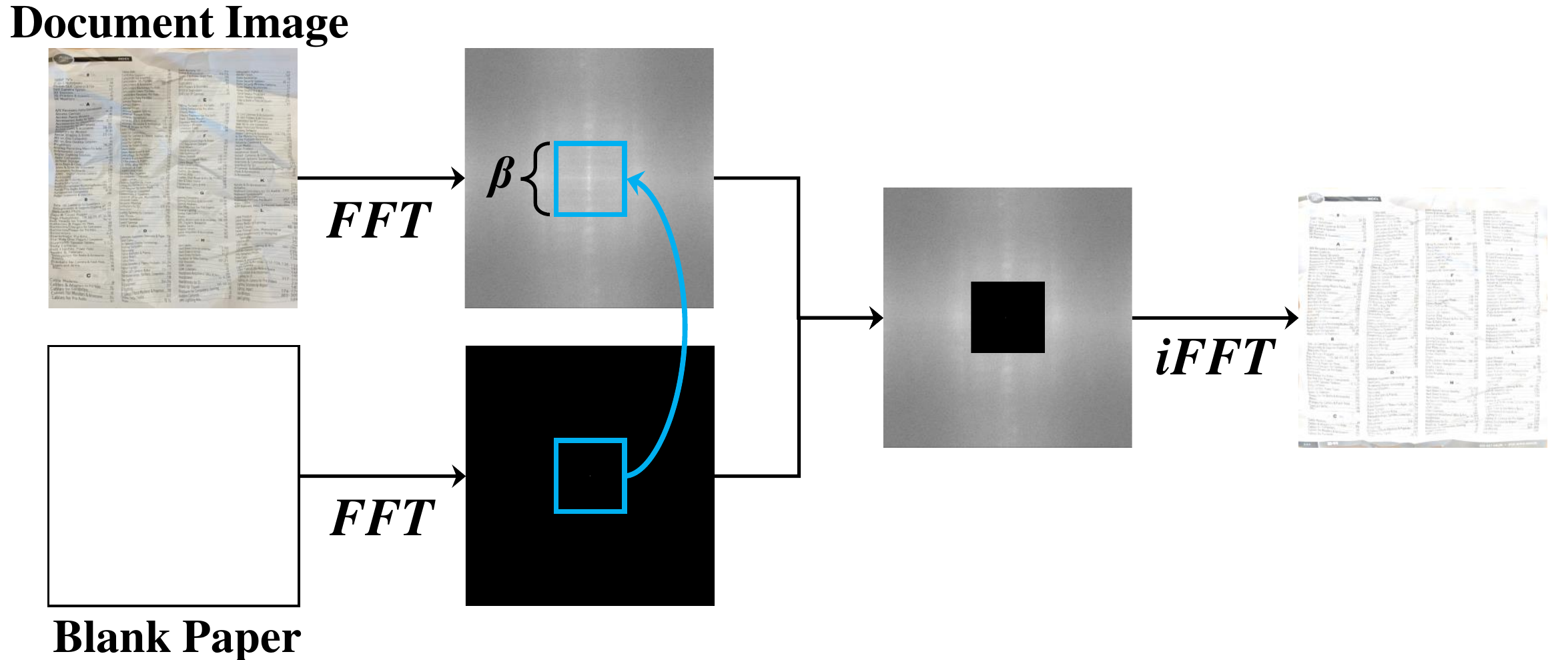}
\caption{
\textbf{Illustration of the proposed Fourier Converter}: The Fourier Converter transforms a document image and a blank paper image into Fourier space by FFT. The low-frequency components of the document image are then replaced by the corresponding components of the blank paper (as highlighted by blue boxes). The modified spectral signals are finally transformed back to spatial space by iFFT, where most low-frequency appearance noises are removed with little effects over high-level content information.
}
\label{fig:fourier}
\end{figure}

\begin{figure*}[!ht]
  \centering
  \includegraphics[width=\linewidth]{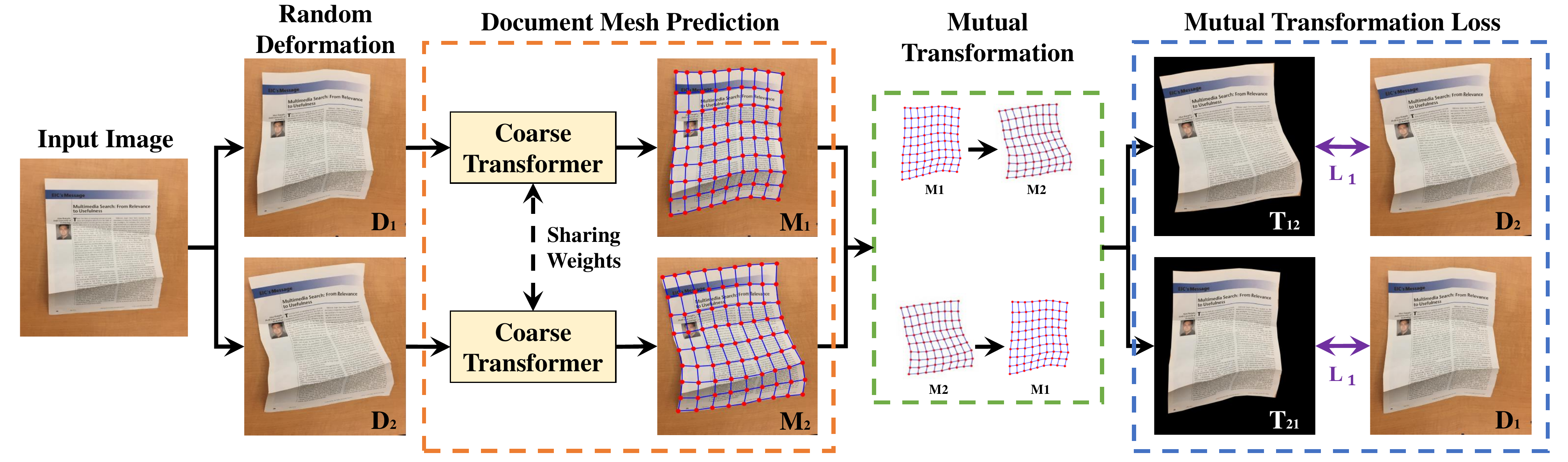}
\caption{\textbf{Illustration of our proposed Mutual Transformation Loss}: Each \textit{Input Image} is randomly deformed to two distorted images $D_1$ and $D_2$. FDRNet then learns to predict mesh grids of the document region $M_1$ and $M_2$ within the two distorted images, and transform them to each other mutually which produces $T_{12}$ and $T_{21}$. The difference between each distorted image and its transformation is computed to guide FDRNet to better focus on document distortion in training.}
\label{fig:mutual}
\end{figure*}

\subsection{Fourier Converter}\label{sec:fourier}

We design a Fourier Converter to extract high-frequency information from document images captured by cameras. Given a \textit{Document Image} as shown in \cref{fig:fourier}, the Fourier Converter first transforms it into Fourier space via Fast Fourier Transform (FFT) \cite{frigo1998fftw}. Next, the document's low-frequency information is replaced with the low-frequency information of a \textit{Blank Paper}. The modified spectral signals are finally transformed back to the spatial space (through inverse Fast Fourier Transform (iFFT)), which produces the OCR-friendly document images with most appearance noises successfully removed.

We employ a hyper-parameter $\beta$ in Fourier Converters in both network training and document recognition tasks. As \cref{fig:fourier} shows, $\beta$ controls how much low-frequency information (the center has the lowest frequency) is replaced. The high-frequency information can thus be extracted with a mask $M_{\beta}$ of size $(H, W)$ as follows:

\begin{equation*}
M_{\beta}(h, w) = \left\{\begin{matrix}
0, & (h, w)\in[-\beta H:\beta H, -\beta W:\beta W]\\ 
1, & Otherwise
\end{matrix} , \right.
\end{equation*}
where $\beta\in[0, 1/2]$, $h\in[-H/2, H/2]$ and $w\in[-W/2, W/2]$. With $x$ denoting the spectral signals of the dewarped document and $x_w$ denoting that of the blank paper, the modified spectral signals can be derived as follows:
\begin{equation}
x' =  M_{\beta} \cdot x + (1 - M_{\beta}) \cdot x_w .
\end{equation}

We empirically set $\beta$ at 0.06 and 0.008, respectively, for the two Fourier Converters in network training and document recognition. Since $\beta$ is a ratio in Fourier space that is invariant to image sizes or resolutions, it can be directly applied to various new images with little tuning, more details to be discussed in the ensuing Experiments.

The Fourier Converter helps to train FDRNet effectively. Given the \textit{Input Image} and the corresponding \textit{Scanned Document}, it first extracts high-frequency information \textit{Input Image (HF)} and \textit{Scanned Document (HF)} as illustrated in \cref{fig:overview}. At each training batch, the \textit{Input Image (HF)} is dewarped to produce \textit{Coarse Dewarping (HF)} and \textit{Refined Dewarping (HF)} by the \textit{Coarse Transformer} and the \textit{Refinement Transformer}, respectively. FDRNet learns by back-propagating $\mathcal{L}_1$ loss between the \textit{Scan Document (HF)} and the \textit{Coarse Dewarping (HF)} \& \textit{Refined Dewarping (HF)}. The Fourier Converter improves network learning from two aspects. First, it discards low-frequency appearance information that often contains rich noisy variations and makes network learning much more complicated. Thanks to such data cleaning, FDRNet can be trained effectively and efficiently by using a small amount of training data. Second, the clear appearance gap between camera-captured documents and scanned documents often affects the stability and convergence of network training. Fourier Converter extracts high-frequency information which minimizes the domain gaps and enables direct loss computation between the two types of document images without requiring any annotations of mesh grids in training.

For document recognition during the inference stage, the Fourier Converter extracts high-frequency information from the \textit{Refined Dewarping} which often suffers from various appearance noises that degrade the document recognition performance greatly. This removes various appearance noises effectively and improves the document recognition greatly as illustrated in \cref{fig:overview}.

\subsection{Network Training}
FDRNet can be trained by optimizing the \textit{Coarse Transformer} and the \textit{Refinement Transformer} only as the \textit{Fourier Converter} is frozen with empirically determined $\beta$ in training. We train the \textit{Coarse Transformer} by using a $L_{rect}$ loss and a mutual transformation loss as follows:
\begin{equation}
\mathcal{L}_{coarse} = \mathcal{L}_{rect} + \lambda * \mathcal{L}_{mutual} ,
\end{equation}
where $\mathcal{L}_{rect}$ is $L_1$ loss that can be directly computed between the \textit{Coarse Dewarping (HF)} and \textit{Scan Document (HF)} as illustrated in \cref{fig:overview}. The $L_1$ loss works well as \textit{Coarse Dewarping (HF)} and \textit{Scan Document (HF)} have similar intensity but little appearance noises and domain gaps. Parameter $\lambda$ is the weight to balance the two losses which is empirically set at 0.5 in our network.

Since document images captured by cameras often suffer from severe geometric distortions, the network training may not converge (with $L_1$ loss alone) without ground-truth annotations of document meshes. We design a mutual transformation loss $L_{mutual}$ that `fabricates' certain supervision to constrain and guide the network to learn geometric distortions stably. The underlying idea of $L_{mutual}$ is that a document with two different geometric distortions can be mutually transformed to each other if their mesh grids are predicted correctly. In implementation, the \textit{Input Image} is first transformed to two new images (i.e. $D_1$ and $D_2$) with randomly perturbed deformation \cite{ma2018docunet} as illustrated in \cref{fig:mutual}. The two transformed images are then fed to FDRNet to predict the corresponding document meshes $M_1$ and $M_2$, respectively. $D_1$ can thus be transformed to image $T_{12}$ by TPS transformation $M_1 \rightarrow M_2$, and $D_2$ can be similarly transformed to image $T_{21}$ by $M_2 \rightarrow M_1$. The mutual transformation loss is thus defined as follows:
\begin{equation}
\mathcal{L}_{mutual} = \|T_{12} - D_2 \| * m_2 + \|T_{21} - D_1 \| * m_1 ,
\end{equation}
where $m_1$ and $m_2$ refer to document regions within $M_1$ and $M_2$. Note although perturbed distortion could produce abnormal distortions around the document background, FDRNet can focus on document regions progressively by restricting loss computation within document meshes and ignoring the document background simultaneously.

The \textit{Refinement Transformer} can be trained by using an $L_1$ loss alone (in between \textit{Refined Dewarping (HF)} and \textit{Scan Document (HF)} as shown in \cref{fig:overview}). It does not require the mutual transportation loss as the \textit{Coarse Transformer} has located document regions and rectified most geometric distortions. The $L_1$ loss alone is sufficient for the prediction of the remaining document distortion. 

\begin{figure*}[!t]
  \centering
  \includegraphics[width=\linewidth]{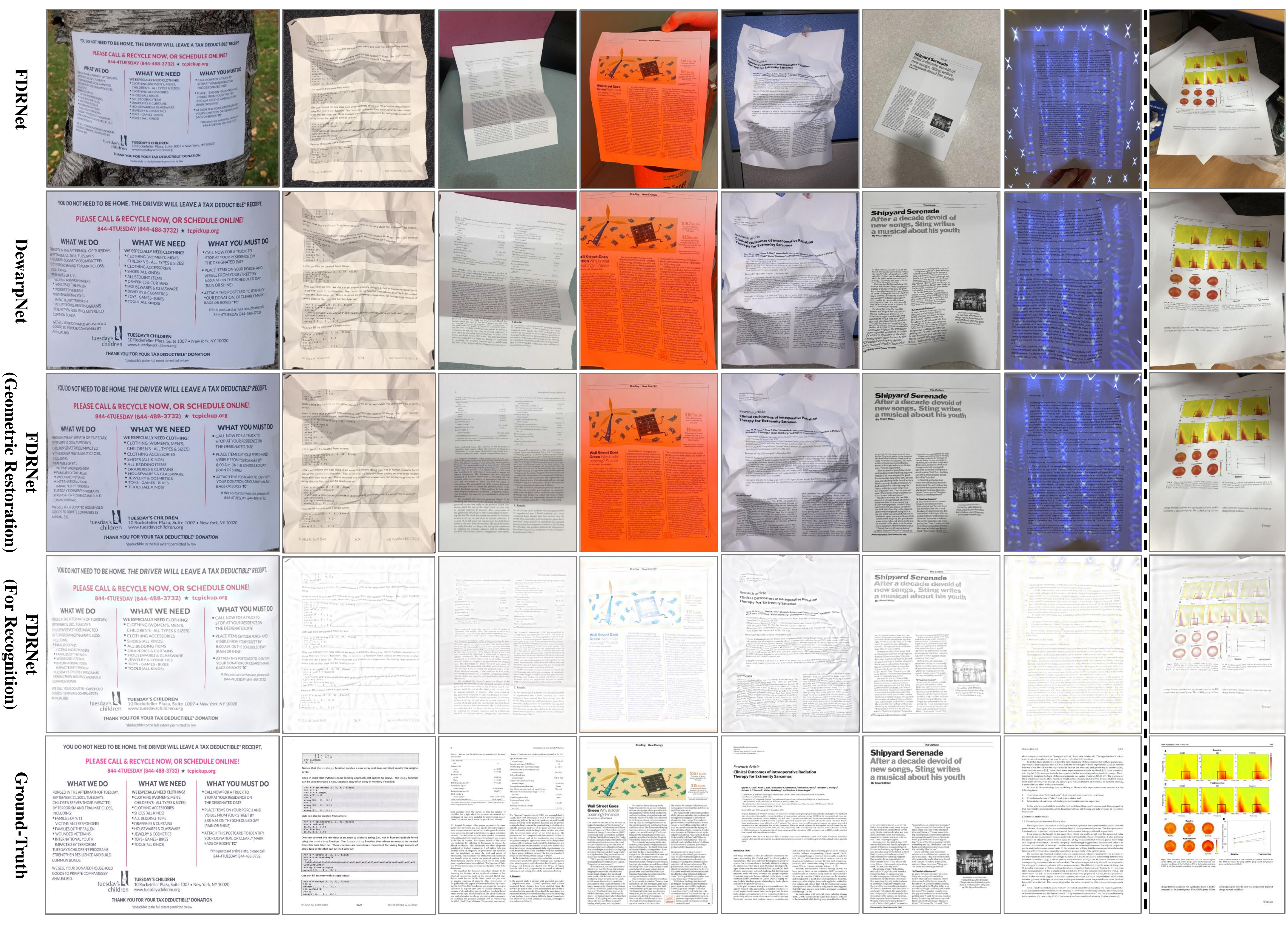}
\caption{\textbf{Illustration of document restoration by FDRNet and DewarpNet}: For the sample images from DocUNet in columns 1-2 and WarpDoc in columns 3-8 in the first row, rows 2 and 3 show the dewarped images by using DewarpNet and FDRNet (dewarping on), respectively. Row 4 shows the appearance restoration by FDRNet which removes various appearance noises and improves document recognition greatly. FDRNet is robust to most geometric and photometric distortions but tends to get confused while document background has similar patterns as document regions as illustrated in the last sample.}
\label{fig:sample}
\end{figure*}

\begin{table}[!t]
\centering
\resizebox{1.0\columnwidth}{!}{
\begin{tabular}{|l|l|}
\hline
\textbf{Deformation} & \textbf{Description}  \\ \hline\hline
Perspective   & With perspective distortion only                        \\ \hline
Fold          & With one or several creases on document                            \\ \hline
Curved        & With curvature distortion                            \\ \hline
Random        & With random crumples                                \\ \hline
Rotating      & With in-plane rotation between -45$\degree$ and 45$\degree$    \\ \hline
\multirow{2}{*}{Incomplete}   
              & With incomplete distortion without affecting\\
              & the content of documents. \\ \hline 
\end{tabular}
}
\caption{\textbf{Details of the proposed WarpDoc Benchmark:} The WarpDoc Benchmark contains 1,020 camera captured document images with six different types of deformation including Perspective, Fold, Curved, Random, Rotating and Incomplete.}
\label{tab:warpdoc}
\end{table}

\begin{table}[!t]
\centering
\resizebox{\columnwidth}{!}{
\begin{tabular}{|l|c|c|c|c|c|}
\hline
\multirow{2}{*}{\textbf{Methods}} & \multicolumn{3}{c|}{\textbf{Training Data}}                                               & \multirow{2}{*}{\textbf{MS-SSIM}} & \multirow{2}{*}{\textbf{LD}} \\ \cline{2-4}
                                  & \textbf{No.} & \multicolumn{1}{c|}{\textbf{D-GT}} & \multicolumn{1}{c|}{\textbf{Type}} &                                   &                              \\ \hline \hline
DocUNet\cite{ma2018docunet}         & 100k & \cmark & Synth                     & 0.41             & 14.08       \\ \hline
GBSUM \cite{bandyopadhyay2020gated} & 8k   & \cmark & Synth                    & 0.42             & 13.20        \\ \hline
AGUN \cite{liu2020geometric}        & 40k  & \cmark & Synth                     & 0.45             & 12.06       \\ \hline
DewarpNet \cite{das2019dewarpnet}   & 100k & \cmark & Synth                     & 0.47             & 8.95        \\ \hline 
\multirow{2}{*}{DocTr \cite{feng2021doctr}}   & 100k & \multirow{2}{*}{\cmark} & Synth                     & \multirow{2}{*}{\textbf{0.50}}         & \multirow{2}{*}{\textbf{8.38}}    \\ 
                                              & +3k &  & +Real                     &          &     \\ \hline 
\textbf{FDRNet}                     & 1k  & \xmark & Real                     &   \textbf{0.50}  &   9.43    \\ \hline
\end{tabular}
}
\caption{\textbf{Image similarity (in MS-SSIM and LD) over DocUNet:} No.: Number of training images; D-GT: Deformation Ground-Truth; Synth: Synthetic images; Real: Real images.}
\label{tab:image_similarity_docu}
\end{table}


\section{Experiments}

\subsection{Datasets}
We evaluated FDRNet over two datasets as listed:

\noindent \textbf{DocUNet \cite{ma2018docunet}:} DocUNet contains 130 images that are taken for different paper documents with different contents and texts of different languages. These images are taken under different conditions which suffer from various distortions. For each paper document, a scanned copy is collected as a ground-truth document.

\noindent \textbf{WarpDoc:} We collected WarpDoc, a warped document image dataset for evaluating document restoration methods. WarpDoc consists of 1,020 camera images of documents that were collected from scientific papers, magazines, envelopes, etc., which have different paper materials, page layouts, and contents. The images were taken in different scenes (indoors, outdoors, etc.) with different illuminations. Before imaging, we warped the 1,020 printed documents into six types of distortions including Fold, Curved, Random, Rotating, Incomplete, and Perspective as illustrated in columns 3-8 of \cref{fig:sample}, respectively. More details about our WarpDoc are available in the Supplementary Material.

\subsection{Evaluation Metrics}
We adopt two types of widely used evaluation metrics \cite{ma2018docunet,das2019dewarpnet,you2017multiview,feng2021doctr} including: 1) Multi-Scale Structural Similarity (MS-SSIM) \cite{wang2003multiscale} and Local Distortion (LD) \cite{you2017multiview} that focus on image similarity performance; 2) Character Error Rate (CER) for evaluation of optical character recognition (OCR) performance. More details are available in the Supplementary Material.

\subsection{Experimental Results}

We conduct cross-validation experiments over DocUNet and WarpDoc benchmarks for evaluation of FDRNet qualitatively and quantitatively. For each test document image, FDRNet model produces two images including a dewarped document image with geometric restoration only and a fully restored document image with further appearance restoration as illustrated in rows 3 and 4 in \cref{fig:sample}, respectively. We evaluate the dewarped document images by using image similarity metrics and the appearance-restored document image by using OCR accuracy.

\begin{table}[!t]
\centering
\resizebox{\columnwidth}{!}{
\begin{tabular}{|l|c|c|c|c|c|}
\hline
\multirow{2}{*}{\textbf{Methods}} & \multicolumn{3}{c|}{\textbf{Training Data}}                                               & \multirow{2}{*}{\textbf{MS-SSIM}} & \multirow{2}{*}{\textbf{LD}} \\ \cline{2-4}
                                  & \textbf{No.} & \multicolumn{1}{c|}{\textbf{D-GT}} & \multicolumn{1}{c|}{\textbf{Type}} &                                   &                              \\ \hline \hline

GBSUM \cite{bandyopadhyay2020gated} & 8k   & \cmark & Synth                    &   0.34         &  29.07  \\ \hline
DewarpNet \cite{das2019dewarpnet}   & 100k & \cmark & Synth                     &  0.33        &  31.15   \\ \hline 
\textbf{FDRNet}                              & 130  & \xmark & Real            & \textbf{0.45}          & \textbf{20.30} \\   \hline\hline
GBSUM-Crop \cite{bandyopadhyay2020gated} & 8k   & \cmark & Synth                    &  0.41           &  23.34  \\ \hline
DewarpNet-Crop \cite{das2019dewarpnet}   & 100k & \cmark & Synth                     & 0.39         &   21.89  \\ \hline 
\textbf{FDRNet-Crop}                              & 130  & \xmark & Real            & \textbf{0.46}          & \textbf{19.11} \\   \hline

\end{tabular}
}
\caption{\textbf{Image similarity (in MS-SSIM and LD) over WarpDoc:} Crop: Evaluation on tightly cropped images from WarpDoc Benchmark; No.: Number of training images; D-GT: Deformation Ground-Truth; Synth: Synthetic images; Real: Real images.}
\label{tab:image_similarity_warpdoc}
\end{table}

\begin{table}[!t]
\centering

\begin{tabular}{|l|c|c|}
\hline
\multirow{2}{*}{\textbf{Methods}}
& \multicolumn{2}{c|}{\textbf{CER(\%)}} \\ 
\cline{2-3}
 & \begin{tabular}[c]{@{}c@{}}DocUNet\\ Benchmark\end{tabular} & \begin{tabular}[c]{@{}c@{}}WarpDoc\\ Benchmark\end{tabular} \\ 
 \hline \hline
 GBSUM \cite{bandyopadhyay2020gated} &  37.94       & 66.48    \\ \hline
DewarpNet\cite{das2019dewarpnet} & 23.95              & 45.82    \\ \hline
DocTr \cite{feng2021doctr} & 20.00 & - \\ \hline
\textbf{FDRNet}                 & \textbf{16.96}       & \textbf{29.24}          \\ \hline
\end{tabular}
\caption{\textbf{Character error rates} over DocUNet and WarpDoc.}
\label{tab:ocr}
\end{table}

\noindent \textbf{Image Similarity:} \cref{tab:image_similarity_docu} shows the MS-SSIM and LD of the proposed FDRNet as well as several state-of-the-art methods over DocUNet and WarpDoc. As \cref{tab:image_similarity_docu} shows, FDRNet achieves competitive dewarping performance over the DocUNet. On the other hand, FDRNet uses much simpler training data than state-of-the-art methods in both image number (1k v.s. 8k-100k) and image annotations (w/o v.s. w/ deformation ground-truth). 

We further evaluate FDRNet on the proposed WarpDoc dataset in which  document images usually suffer from much more complex distortions than document images in DocUNet benchmark. We conduct two sets of experiments for a better comparison with the state-of-the-art. First, we compare FDRNet with existing document dewarping methods on the original WarpDoc dataset to evaluate document dewarping under the presence of both complex geometric distortions and significant depth variation. Second, we crop the images in the WarpDoc dataset (following \cite{ma2018docunet}) to reduce the depth variation of documents in the original images. We hence compare FDRNet with existing methods on the cropped images in which depth variations of documents are largely mitigated. As \cref{tab:image_similarity_warpdoc} shows, the proposed FDRNet outperforms the existing approaches on dewarping documents with complex geometric distortions alone or additional depth variation by using much fewer and simpler training samples. This result shows that the proposed FDRNet is more robust to document dewarping as compared with the state-of-the-art. Additionally, the GBSUM and DewarpNet achieve very different performances on dewarping original and cropped document images, showing that they are sensitive to document depth variations. On the contrary, the dewarping performance of FDRNet on original and cropped images is similar, demonstrating the proposed FDRNet is much more robust to the depth variation of documents as compared with existing approaches.

\cref{fig:sample} shows the restoration of several sample images from DocUNet and WarpDoc that suffer from different types of distortions. As \cref{fig:sample} shows, FDRNet achieves similar restoration as DewarpNet for documents with simple curvature distortions (sample in column 1). But for documents with more complex distortions in columns 2-7, FDRNet usually performs better as it focuses on high-frequency information where document layouts such as text lines help to learn geometric distortions better. As a comparison, 3D methods such as DewarpNet regress each pixel from warped documents to flat ones. The regression of pixels around complex crumples is often hard to learn as there are much fewer such pixels as compared with those with simple distortions. FDRNet learns a general transformation by a coarse mesh grid which is less affected by the pixel-level data imbalance during network training.

\begin{table}[!t]
\centering
\resizebox{\columnwidth}{!}{
\begin{tabular}{|c|c|c|c|c|c|c|c|}
\hline
 \multicolumn{5}{|c|}{\textbf{FDRNet Components}}         & \multicolumn{3}{c|}{\textbf{Experimental Results}}        \\ \hline
  \textbf{\textit{CT}} & \textbf{\textit{FC$_{tr}$}} & \textbf{\textit{MTL}} & \textbf{\textit{RT}} & \textbf{\textit{FC$_{inf}$}} & \textbf{MS-SSIM} & \textbf{LD}      & \textbf{CER(\%)} \\ \hline\hline
\cmark        &            &              &           &           &  \multicolumn{3}{c|}{Not converge} \\ \hline
\cmark        & \cmark     &              &           &           &  0.32            & 34.16            &  69.32  \\ \hline
\cmark        & \cmark     &              & \cmark    &           &  0.37            & 23.47            &  48.24   \\ \hline
\cmark        & \cmark     & \cmark       &           &           &  0.44            & 16.35            &  33.02       \\ \hline
\cmark        & \cmark     & \cmark       & \cmark    &           &  \textbf{0.50}   & \textbf{9.43}   &  23.46   \\ \hline
\cmark        & \cmark     & \cmark       & \cmark    & \cmark    &  -               &  -               &  \textbf{16.96}      \\ \hline
\end{tabular}
}
\caption{\textbf{Ablation study} of FDRNet over DocUNet: \textit{FC$_{tr}$} - Fourier Converter for training; \textit{CT} - Coarse Transformer; \textit{MTL} - Mutual Transformation Loss; \textit{RT} - Refinement Transformer; \textit{FC$_{inf}$} - Fourier Converter for inference.}
\label{tab:ablation}
\end{table}

\noindent \textbf{OCR Performances: } We examine how FDRNet performs on document recognition by evaluating OCR over FDRNet restored documents using PyTesseract (v4.1.1) \cite{smith2007overview}. Following DewarpNet \cite{das2019dewarpnet}, we perform OCR over 54 document images on DocUNet and 739 document images on WarpDoc with lots of texts. 
\cref{tab:ocr} shows experimental results. We can observe that FDRNet achieves CER of 16.96\% and 29.24\% on DocUNet and WarpDoc, respectively, which outperforms state-of-the-art methods with illumination restoration by large margins. More specifically, although the performances of FDRNet and state-of-the-art methods are comparable on the metrics of image similarity on DocUNet dataset as shown in \cref{tab:image_similarity_docu}, FDRNet outperforms these approaches by a large margin on CER, demonstrating that FDRNet is more robust to document recognition task. The second last row in \cref{fig:sample} illustrates the FDRNet restored documents images. It can be seen that FDRNet removes various geometric and appearance distortions from the dewarped documents which facilitate OCR and document recognition significantly.

\subsection{Discussion}
    
\noindent\textbf{Ablation studies:} We study the contributions of different designs in our FDRNet including a Fourier Converter for network training \textit{FC$_{tr}$}, a Coarse Transformer \textit{CT}, a Mutual Transformation Loss \textit{MTL}, a Refinement Transformer \textit{RT} and a Fourier Converter for inference \textit{FC$_{inf}$}. \cref{tab:ablation} shows the experimental results.

As shown in \cref{tab:ablation} rows 1-2, the \textit{CT} alone cannot converge due to unstable losses during training that are caused by the large domain gap between document images collected by cameras and scanners. By including the proposed \textit{FC$_{tr}$}, FDRNet training stabilizes. The further inclusion of \textit{MTL} and \textit{RT} both help to train more powerful dewarping models with clearly improved MS-SSIM and LD, as shown in rows 3-5. During inference, including the Fourier Converter (i.e. \textit{FC$_{inf}$}) improves OCR by large margins as \textit{FC$_{inf}$} removes various appearance noises that often affect document recognition, as shown in row 6.

\begin{table}[!t]
\centering
\begin{tabular}{|c|c|c|c|c|c|}
\hline
$\beta$  & 0.003 & 0.005 & 0.008 & 0.01  & 0.02  \\ \hline
\textbf{CER(\%)}  & 18.52 & 17.84 & 16.96 & 17.38 & 17.72 \\ \hline
\end{tabular}
\caption{
CER varies with the parameter $\beta$ in the Fourier Converter (described in Section 3.2 and Fig. 4.).
}
\label{tab:beta}
\end{table}

\noindent \textbf{Parameter $\beta$:} Parameter $\beta$ in the Fourier Converter (\cref{sec:fourier}) affects FDRNet at both network training and document recognition (inference) stages. Specifically, FDRNet converges well when $\beta$ lies within a suitable range. In addition, FDRNet recognition is not sensitive to $\beta$ either. As \cref{tab:beta} shows, the CER of the trained FDRNet models is quite stable when $\beta$ changes in certain ranges. More details about parameter $\beta$ on model training are provided in the Supplementary Material.

\noindent \textbf{Constraints:} The proposed FDRNet may be confused if the document background region has similar patterns as the document region. Under such situations, document background could be treated as parts of document region in restoration as illustrated in the last sample in \cref{fig:sample}.


\section{Conclusion and Future Work}
This paper presents a document restoration network FDRNet for better recognition of document images captured by cameras. FDRNet focuses on high-frequency information in the Fourier space which allows it to learn from a small amount of training data effectively. Additionally, FDRNet can generalize to new data well as it discards low-frequency information which mitigates domain gaps greatly. Extensive experiments show that FDRNet is capable of removing geometric and appearance degradation which improves document recognition significantly. In the future, we would like to study the simple yet effective image synthesis and so to leverage the advances of training on both real and synthetic data for more robust document dewarping and recognition.

{\small
\bibliographystyle{ieee_fullname}
\bibliography{egbib}
}

\end{document}